\documentclass{article} % For LaTeX2e
\usepackage{iclr2019_conference,times}

\usepackage{amsmath,amsfonts,bm}

\def\eqref#1{equation~\ref{#1}}
\def\1{\bm{1}}

\DeclareMathAlphabet{\mathsfit}{\encodingdefault}{\sfdefault}{m}{sl}
\SetMathAlphabet{\mathsfit}{bold}{\encodingdefault}{\sfdefault}{bx}{n}

\usepackage{hyperref}
\usepackage{url}

\usepackage[subtle]{savetrees}
\usepackage{graphicx}
\usepackage{subcaption}
\usepackage{mathtools}

\newcommand{\minihead}[1]{{\vspace{.45em}\noindent\textbf{#1.} }}

\title{LIT: Block-wise Intermediate Representation Training for Model Compression}

\author{Animesh Koratana*, Daniel Kang*, Peter Bailis, Matei Zaharia}

\iclrpreprintcopy
\begin{document}

\maketitle

\begin{abstract}

Knowledge distillation (KD) is a popular method for reducing the computational
overhead of deep network inference, in which the output of a teacher model is
used to train a smaller, faster student model. Hint training (i.e., FitNets)
extends KD by regressing a student model's intermediate representation to a
teacher model's intermediate representation. In this work, we introduce
b\textbf{L}ock-wise \textbf{I}ntermediate representation \textbf{T}raining (LIT), a novel model compression technique that
extends the use of intermediate representations in deep network compression,
outperforming KD and hint training. LIT has two key ideas: 1) LIT trains a student of
the same width (but shallower depth) as the teacher by directly comparing the
intermediate representations, and 2) LIT uses the intermediate representation
from the previous block in the teacher model as an input to the current student
block during training, avoiding unstable intermediate representations in the
student network. We show that LIT provides substantial reductions in network
depth \emph{without loss in accuracy} --- for example, LIT can compress a ResNeXt-110
to a ResNeXt-20 (5.5$\times$) on CIFAR10 and a VDCNN-29 to a VDCNN-9
(3.2$\times$) on Amazon Reviews without loss in accuracy, outperforming KD and
hint training in network size for a given accuracy. We also show that applying LIT to
identical student/teacher architectures increases the accuracy of the student
model above the teacher model, outperforming the recently-proposed Born Again
Networks procedure on ResNet, ResNeXt, and VDCNN.  Finally, we show that LIT can
effectively compress GAN generators, which are not supported in the KD framework
because GANs output pixels as opposed to probabilities.

\end{abstract}

\section{Introduction}

Modern deep networks have achieved increased accuracy by continuing to introduce
more layers~\citep{ioffe2015batch, he2016deep} at the cost of higher
computational overhead. In response, researchers have proposed many techniques
to reduce this computational overhead at inference time, which broadly fall
under two categories. First, in deep compression~\citep{han2015deep,
zhu2016trained, li2016pruning, hubara2017quantized}, parts of a model are
removed or quantized to reduce the number of weights and/or the computational
footprint.\footnote{In this work, we refer to this class of methods as ``deep
compression,'' and methods to reduce model size more generally as ``model
compression.''} However, deep compression techniques typically require new
hardware~\citep{han2016eie} to take advantage of the resulting model sparsity.
Second, in student/teacher methods---introduced in knowledge distillation
(KD)~\citep{hinton2015distilling} and further extended~\citep{romero2014fitnets,
kim2016sequence, furlanello2018born}---a smaller student model learns from a
large teacher model through distillation loss, wherein the student model
attempts to match the logits of the teacher model. As there are no constraints
on the teacher and student models, KD can produce hardware-friendly models: the
student can be a standard model architecture (e.g., ResNet), optimized for a
given hardware substrate.

Hint training (i.e., FitNets~\citep{romero2014fitnets}) extends KD by using a
teacher's intermediate representation (IR, i.e., the output from a hidden layer)
to guide the training of the student model. The authors show that hint training
with a single IR outperforms KD in compressing teacher networks (e.g., maxout
networks~\citep{goodfellow2013maxout}) to thinner and deeper student networks.

We ask the natural question: does hint training compress more modern,
highly-structured, very deep networks---such as
ResNet~\citep{he2016deep}, ResNeXt~\citep{xie2017aggregated},
VDCNN~\citep{conneau2016very}, and StarGAN~\citep{choi2017stargan}? We
find that standard hint training (i.e., with a single hint) and
training with multiple hints is not effective for modern deep networks
(Section~\ref{sec:eval-training}). We hypothesize that, for modern
deep networks, hint training causes unstable IRs: the deepest network
considered in~\cite{romero2014fitnets} was only 17 layers, achieving
91.61\% on CIFAR10; in contrast, a modern 110-layer ResNet achieves
93.68\% on CIFAR10.
%These modern network architectures frequently
%consist of repeated blocks (i.e., groups of layers) that can be scaled
%up and down to achieve accuracy/speed trade-offs; for example, ResNets
%have standard configurations from 20 to hundreds of layers.

\begin{figure}[t!]
  \centering
  \includegraphics[width=0.95\columnwidth]{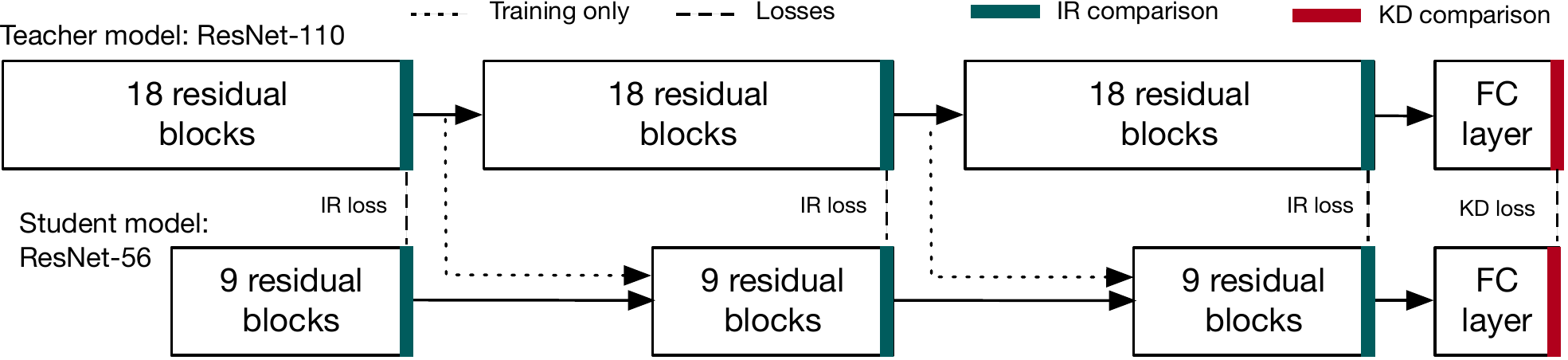}
  \vspace{-0.3em}
  \caption{A schematic of LIT. In LIT, the teacher model's blocks are used as
  input to the student model's blocks during training, except for the first
  block. Specifically, denoting the blocks $S_1, ..., S_4$ for the student and
  $T_1, ..., T_4$ for the student, $S_2(T_1)$ is compared against $T_2$ in
  training and similarly for deeper parts of the network. $S_1$ and $T_1$ are
  directly compared. LIT additionally compares $S$ and $T$ through the KD loss.
  The teacher model is not updated in training.}
  \vspace{-0.7em}
  \label{fig:lit-arch}
\end{figure}

In this work, we extend hint training's ability to transfer intermediate
knowledge from teacher to student to reduce the depth of modern,
highly-structured architectures (e.g., compressing a standard ResNeXt-110 to a
standard ResNeXt-20 with no loss in accuracy). We do this via a novel method
called b\textbf{L}ock-wise \textbf{I}ntermediate representation
\textbf{T}raining (LIT), a student/teacher compression technique that
outperforms training student networks from scratch, hint training, and KD. LIT
targets highly structured, modern networks that consist of repetitive blocks
(i.e., groups of layers) that can be scaled up/down for accuracy/speed
trade-offs; for example, ResNets have standard configurations from 20 to
hundreds of layers. LIT leverages two key ideas to reduce unstable IRs in deep
networks. First, LIT directly
trains student networks of the \emph{same width} as the teacher model
(as opposed to using a single, thinner hint as in hint training). Second,
LIT avoids unstable student IRs deep in the network by using the IR
from the \emph{previous} block in the teacher model as input to the
current student block during training; each student block is
effectively trained in isolation to match the corresponding (deeper)
block in the teacher. We show that LIT's block-wise training improves
accuracy, allows for copying parts of the teacher model directly to
the student model, and permits selective compression of networks
(e.g., compressing one out of three blocks in a network and copying
the rest). For example, consider compressing a ResNet-56 from a
ResNet-110 (Figure~\ref{fig:lit-arch}), each of which have four
sections. The IR loss is applied to the output of each block, and the
teacher model's IRs are used as input to the student blocks.

Because it is possible to transfer IRs directly, LIT is, to our knowledge, the
first student/teacher compression method that works for GAN generators. LIT can
compress GAN generators by only compressing the repetitive blocks present in
certain GANs~\citep{choi2017stargan}. In contrast, KD does not apply directly as the KL divergence in
KD loss operates on probabilities but not the pixels output by GAN generators. LIT
can compress GANs by leveraging LIT's key property that, by matching the teacher
IR dimensions, parts of the teacher network can be directly copied to the
student network.

We show that LIT outperforms standard KD on a range of models (ResNet, ResNeXt,
VDCNN, StarGAN) and datasets (CIFAR10, CIFAR100, Amazon Reviews, CelebA):
empirically, LIT can reduce model sizes from 1.7$\times$ to 5.5$\times$ with no
loss in accuracy. Recent work on Born Again networks~\citep{furlanello2018born}
uses standard KD to train identical student and teacher models to higher
accuracies (i.e., no compression). We show that the benefits of this procedure
also apply to LIT student/teacher training, and LIT enables up to 0.64\% higher
accuracy than KD-based Born Again networks on the networks we consider.

\section{Related Work}
\label{sec:related-work}

\minihead{Knowledge distillation}
\cite{hinton2015distilling, bucilua2006model} introduced knowledge distillation
(KD) in which a teacher ensemble or model’s outputs are used to train a smaller
student model, which inspired a variety of related methods, e.g., for
cross-modal distillation or faster training~\citep{gupta2016cross,
chen2015net2net, frosst2017distilling, romero2014fitnets, furlanello2018born}.
FitNets extends KD by regressing a student model's IR to a teacher model's IR,
as the student models they consider are thinner and deeper.
\cite{wang2018progressive} extends FitNets by training networks iteratively
using hints. In contrast, LIT uses the teacher IRs as input to the student model
in training and directly penalizes deviations of the student model's IRs from
teacher model's IRs, which helps guide training for higher accuracy and improved
inference performance. In Born Again networks~\citep{furlanello2018born}, the
same network architecture is used as both the teacher and student in standard
KD, resulting in higher accuracy. We show that LIT outperforms the Born Again
procedure on ResNet, ResNeXt, and VDCNN.

\minihead{Deep compression}
In deep compression, parts of a network (weights, groups of weights, kernels, or
filters~\citep{mao2017exploring}) are removed for efficient
inference~\citep{han2015deep}, and the weights of the network are quantized,
hashed, or compressed~\citep{hubara2016binarized, rastegari2016xnor,
zhu2016trained, hubara2017quantized}. These methods largely do not take
advantage of a teacher model and typically require new hardware for efficiency
gains~\citep{han2016eie}. Methods that prune filters~\citep{li2016pruning} can
result in speedups on existing hardware, but largely degrade accuracy. We show
that LIT models can be pruned, and thus these methods are complementary to LIT.

\minihead{Network architectures for fast inference}
Researchers have proposed network architectures (e.g.,
MobileNet~\citep{howard2017mobilenets}) and new operations for fast inference
(e.g., ShuffleNet~\citep{zhang2017shufflenet}) on specific hardware. However,
these architectures and operations are largely designed for
power/resource-constrained mobile devices and sacrifice accuracy for low power.
We focus on highly accurate, very deep networks in this work.

\section{Methods}
\label{sec:methods}

LIT uses an augmented loss function and training procedure to distill a teacher
model into a student model. In its training procedure, LIT both 1)
penalizes deviations of the student model's IRs from the teacher model's IRs (IR
loss) and 2) uses the KD loss (for the entire student network). As LIT directly penalizes deviations in IRs, LIT
requires that the teacher model and student model have outputs of the same size
at some intermediate layer.

A key challenge in the LIT procedure is that the student network will not have
meaningful IRs for a large part of the training (e.g.,
at the start of training when the weights are initialized randomly). To address
this issue, LIT uses the teacher model's IRs as inputs
to the student model (described below).

We describe the overall LIT procedure, describe the KD loss, describe how IRs
are used in LIT, and discuss hyperparameter optimization.

\minihead{LIT}
In LIT, we combine the KD and IR loss. We
show that combining the losses results in smaller models for a fixed accuracy in
Section~\ref{sec:experiments}. Specifically, for teacher $T$ and student $S$ the full LIT loss is:
\begin{equation}
  \beta \cdot \mathcal{L}_{\mathrm{KD},\alpha}(T, S) +
  (1 - \beta) \cdot \mathcal{L}_{\mathrm{I}}(T, S)
\end{equation}
with $\alpha,\beta \in [0, 1]$ ($\alpha$ is described below, $\beta$ is an
interpolation parameter). In some cases, we use $\beta = 0$, i.e., we only use
the IR loss (e.g., for GANs, where KD does not apply).

As the IRs have matching dimensions, LIT also allows
parts of the teacher model to be copied directly into the student model. For
example, for ResNets, we copy the teacher's first convolution (before the skip
connections) and fully connected layer to the student model. LIT can also be
used to compress specific parts of a model, as we do with StarGAN's
generator~\citep{choi2017stargan}.

Finally, after we train the student model with LIT, we fine-tune the student
model with the KD loss.

\minihead{Knowledge distillation loss}
In KD, a (typically larger) teacher model or ensemble is used to train a
(typically smaller) student model. Specifically, the KL-divergence between the
probabilities of the student and teacher model is minimized, in addition to the
standard cross-entropy loss.

Formally, denote (for the teacher model) $q_i^\tau = \tfrac{\exp(z_i /
\tau)}{\sum_j z_j / \tau}$ where $z_i$ are the inputs to the softmax and $\tau$
is a hyperparmeter that ``softens" the distribution. Denote $p_i^\tau$ to be the
corresponding quantity for the student model.

Then, the full KD loss is:
\begin{equation}
  \mathcal{L}_{\mathrm{KD}}(p, q, y) = \alpha \cdot H(y, p) + (1 - \alpha) \cdot H(p, q)
\end{equation}
for $y$ to be the true labels, $H$ to be the cross-entropy loss, and $\alpha$ to
be the interpolation parameter.

\cite{hinton2015distilling} sets $\alpha = 0.5$, but we show that the choice of
$\alpha$ can affect performance
(Section~\ref{sec:hyperparameters}).

\minihead{Training via intermediate representations}
In LIT, we logically divide the student and teacher networks into $k$
sub-networks such that the input and output dimensions match for the
corresponding sub-networks (an example is shown in Figure~\ref{fig:lit-arch}).

Denote the full teacher network and student network to be $T$ and $S$
respectively. Denote the teacher sub-networks to be $T_i$ and the student
sub-networks to be $S_i$ such that $T_i, S_i : \mathbb{R}^{n_{i-i}} \to
\mathbb{R}^{n_{i}}$ and that the composition of the sub-networks is the full
network, namely that $T_k(T_{k-1} (\cdots T_1(x))) = T(x)$. We will omit the
argument when convenient.

Denote the loss on the IR loss $l$ (e.g., L2 loss). The
full intermediate loss (given the set of splits) is:
\begin{equation}
  \mathcal{L}_{\mathrm{I}}(T, S) \vcentcolon=
  l(S_1, T_1) + 
  \textstyle \sum_{i=2}^{k} l(S_i(T_{i-1}), T_i)
\end{equation}

Concretely, consider a ResNet-110 as the teacher and a ResNet-56 as the student,
each with three ``sections", i.e., layers in the network with downsampling, and
an L2 intermediate loss. Here, the first teacher ResNet ``section" is
$T_1$, etc. and the L2 deviation from the feature maps, across all the
downsampling feature maps, is the full intermediate loss. A schematic is shown
in Figure~\ref{fig:lit-arch}.

This procedure has two key decisions: 1) where to logically split the teacher
and student models, and 2) the choice of IR loss.
We discuss these settings in the hyperparameter optimization below.

\minihead{Hyperparameter optimization}
LIT inherits two hyperparameters from KD and introduces one more: $\tau$
(the temperature in KD), $\alpha$ (the interpolation parameter in KD), and
$\beta$ (the interpolation parameter in LIT), along with an intermediate
representation loss and split points. In this work, we only consider adding the
IR loss between natural split points, e.g., when a downsampling occurs in a
convolutional network. We have additionally
found that L2 loss works well in practice, so we use the L2 loss for all
experiments unless otherwise noted (Section~\ref{sec:hyperparameters}).

We have found that iteratively setting $\tau$, then $\alpha$, then $\beta$ to
work well in practice. We have found that the same hyperparameters work well for
a given student and teacher structure (e.g., ResNet teacher and ResNet student).
Thus, we use the same set of hyperparameters for a given student and teacher
structure (e.g., we use the same hyperparameters for a teacher/student of
ResNet-110/ResNet-20 and ResNet-110/ResNet-32). To set the hyperparameters for a
given structure, we first set $\tau$ using a small student model, then $\alpha$
for the fixed $\tau$, then $\beta$ for the fixed $\alpha$ and $\tau$ (all on the
validation set).

\section{Experiments}
\label{sec:experiments}

We evaluate LIT's efficacy at compressing models on a range of tasks and models,
including image classification, sentiment analysis, and image-to-image
translation (GAN). Throughout, we use student and teacher networks with the same
broad architecture (e.g., ResNet to ResNet). We consider
ResNet~\citep{he2016deep}, ResNeXt~\citep{xie2017aggregated},
VDCNN~\citep{conneau2016very}, and StarGAN~\citep{choi2017stargan}. We use
standard architecture depths, widths, and learning rate schedules, and perform
hyperparameter selection for the KD and LIT interpolation parameters, while also
performing a sensitivity analysis of these hyperparameters in the sequel.

\subsection{LIT significantly compresses models}

\begin{table}[t!]
\centering
\small
\setlength\itemsep{2em}
\begin{tabular}{lll}
  Dataset        & Task                                & Models          \\ \hline
  CIFAR10        & Image classification                & ResNet, ResNeXt \\
  CIFAR100       & Image classification                & ResNet, ResNeXt \\
  Amazon Reviews & Sentiment analysis (full, polarity) & VDCNN
\end{tabular}
\vspace{-0.5em}
\caption{List of datasets, tasks, and models for standard classification tasks
we compress with LIT. We additionally compress StarGAN on CelebA.}
\label{table:dataset-model}
\vspace{-0.5em}
\end{table}

\minihead{LIT is effective at compressing a range of datasets and models}
We ran LIT on a variety of models and datasets for image classification and
sentiment analysis (Table~\ref{table:dataset-model}). We additionally performed
KD and hint training on these datasets and models. We selected the hyperparameters sequentially
(Section~\ref{sec:methods}).

\begin{figure}[t!]
  \centering
  \begin{subfigure}{.49\columnwidth}
    \includegraphics{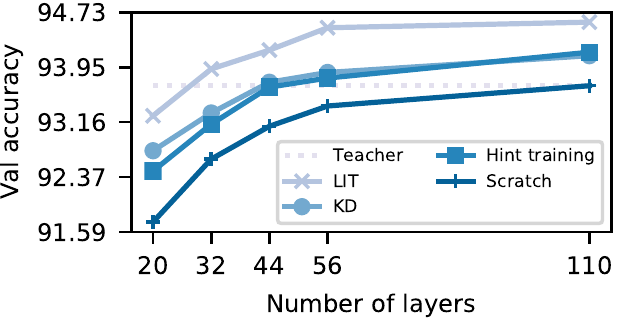}
    \vspace{-0.5em}
    \caption{CIFAR10, ResNet, end-to-end accuracy}
  \end{subfigure}
  \begin{subfigure}{.49\columnwidth}
    \includegraphics{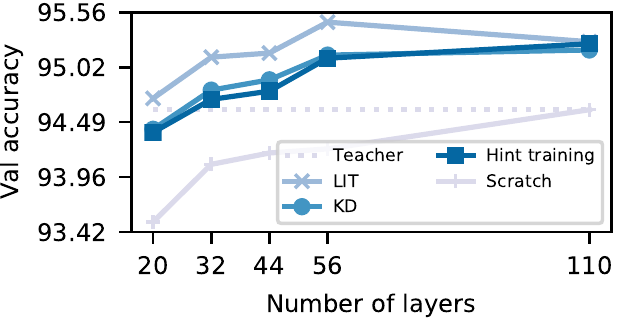}
    \vspace{-0.5em}
    \caption{CIFAR10, ResNeXt, end-to-end accuracy}
  \end{subfigure}

  \begin{subfigure}{.49\columnwidth}
    \includegraphics{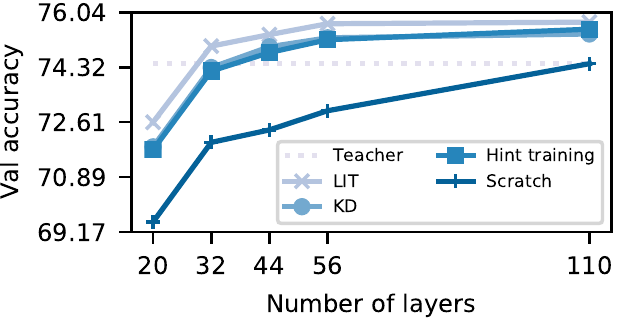}
    \vspace{-0.5em}
    \caption{CIFAR100, ResNet, end-to-end accuracy}
  \end{subfigure}
  \begin{subfigure}{.49\columnwidth}
    \includegraphics{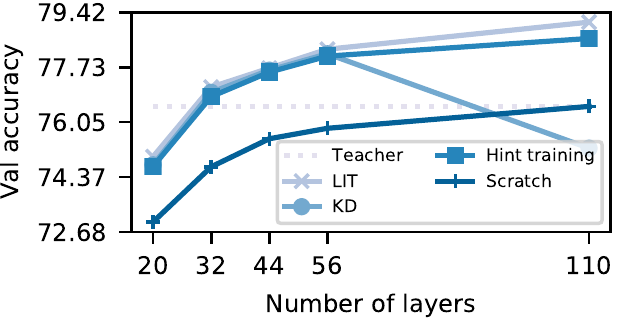}
    \vspace{-0.5em}
    \caption{CIFAR100, ResNeXt, end-to-end accuracy}
  \end{subfigure}

  \vspace{-0.5em}
  \caption{The accuracy of ResNet and ResNeXt trained from scratch, trained via
  KD, and trained via LIT for CIFAR10/100. The teacher model was ResNet-110 and
  ResNeXt-110 respectively. As shown, LIT outperforms KD for every student
  model. The student architecture being the same as the parent architecture
  corresponds to born again networks, which LIT also outperforms. In some cases,
  KD can reduce the accuracy of the student model, as reported
  in~\cite{mishra2017apprentice}.}
  \label{fig:cifar10}
  \vspace{-0.5em}
\end{figure}

\begin{figure}[t!]
  \centering
  \begin{subfigure}{.49\columnwidth}
    \includegraphics{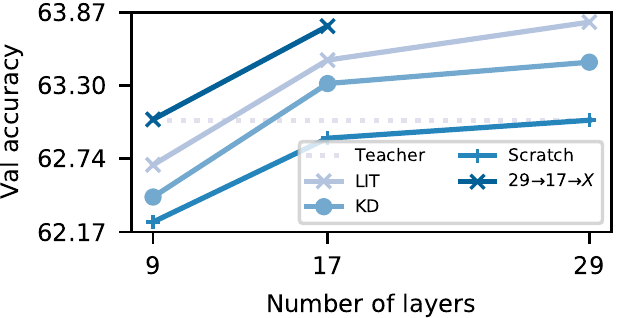}
    \vspace{-0.5em}
    \caption{Amazon full, VDCNN, end-to-end accuracy}
  \end{subfigure}
  \begin{subfigure}{.49\columnwidth}
    \includegraphics{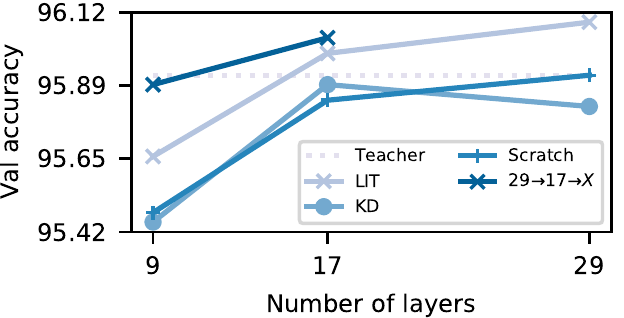}
    \vspace{-0.5em}
    \caption{Amazon polarity, VDCNN, end-to-end accuracy}
  \end{subfigure}
  \vspace{-0.5em}
  \caption{The accuracy of VDCNN on Amazon reviews (full and polarity) trained
  from scratch, trained via KD, and trained via LIT.}
  \label{fig:vdcnn}
  \vspace{-0.5em}
\end{figure}

Figure~\ref{fig:cifar10} shows the results for ResNet and ResNeXt for
CIFAR10 and CIFAR100, and Figure~\ref{fig:vdcnn} shows the results for
VDCNN on Amazon Reviews (full, polarity). LIT can compress
models by up to 5.5$\times$ (CIFAR10, ResNeXt 110 to 20) on image
classification and up to 3.2$\times$ on sentiment analysis (Amazon
Reviews, VDCNN 29 to 9), with no loss in accuracy. LIT outperforms KD and hint
training
on all settings, resulting in up to 5.5$\times$ smaller models with no
loss in accuracy (CIFAR10, ResNeXt-110 vs ResNeXt-20). Additionally, LIT
outperforms the recently proposed Born Again procedure in which the same
architecture is used as both the student and teacher
model~\citep{furlanello2018born} (i.e., only for improved accuracy, not
for compression). We have found that, in some
cases, training sequences of models using LIT results in higher
performance.  Thus, for VDCNN, we additionally compressed using LIT a
VDCNN-29 to a VDCNN-17, and using this VDCNN-17, we trained a VDCNN-9
and VDCNN-17.

Importantly, LIT can retain the same architectural patterns for both the student
and teacher model and does not require that models be deeper and thinner as in
FitNets~\citep{romero2014fitnets}.

We also found that in some cases, KD degrades the accuracy of student models
when the teacher model is the same architecture (ResNeXt-110 on
CIFAR100, VDCNN-29 on Amazon Reviews polarity). This corroborates prior
observations in~\cite{mishra2017apprentice}.

%\begin{figure}[t!]
%  \centering
%  \begin{subfigure}{.49\columnwidth}
%    \includegraphics{figures/fu-cifar100-resnet.pdf}
%    \vspace{-0.5em}
%    \caption{CIFAR100, ResNet, end-to-end accuracy}
%  \end{subfigure}
%  \begin{subfigure}{.49\columnwidth}
%    \includegraphics{figures/fu-cifar100-resnext.pdf}
%    \vspace{-0.5em}
%    \caption{CIFAR100, ResNeXt, end-to-end accuracy}
%  \end{subfigure}
%  \vspace{-0.5em}
%  \caption{The accuracy of ResNet and ResNeXt trained from scratch, trained via
%  KD, and trained via LIT for CIFAR100. In some cases, KD can reduce the
%  accuracy of the student model, as reported in~\cite{mishra2017apprentice}.}
%  \label{fig:cifar100}
%  \vspace{-0.5em}
%\end{figure}

\minihead{LIT can reduce group cardinality}
While LIT requires the size of at least one IR to be the same width between the
teacher and student model, several classes of models have an internal width or
group \emph{cardinality}. For example, ResNeXt~\citep{xie2017aggregated} has a ``grouped
convolution", which is equivalent to several convolutions with the same input
(see Figure $3$ in~\cite{xie2017aggregated}). The width of the network is not
affected by the group size, so LIT is oblivious to the group size.

\begin{figure}[t!]
  \centering
  \includegraphics{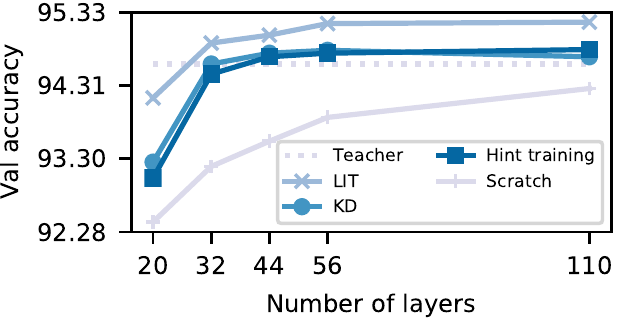}
  \vspace{-0.5em}
  \caption{ResNeXt student models with cardinality 16 trained from a ResNeXt-110
  with cardinality 32. We show that LIT can reduce the cardinality and that LIT
  outperforms KD.}
  \label{fig:cifar10-resnext-card}
  \vspace{-0.5em}
\end{figure}

We show that LIT can reduce the group cardinality for ResNeXt. We
train student ResNeXts with cardinality 16 (instead of the default 32)
from a ResNeXt-110 (cardinality
32). Figure~\ref{fig:cifar10-resnext-card} illustrates the
results. As before, LIT outperforms KD and training from scratch in
this setting.

\begin{figure}[t!]
  \centering
  \includegraphics[width=0.9\columnwidth]{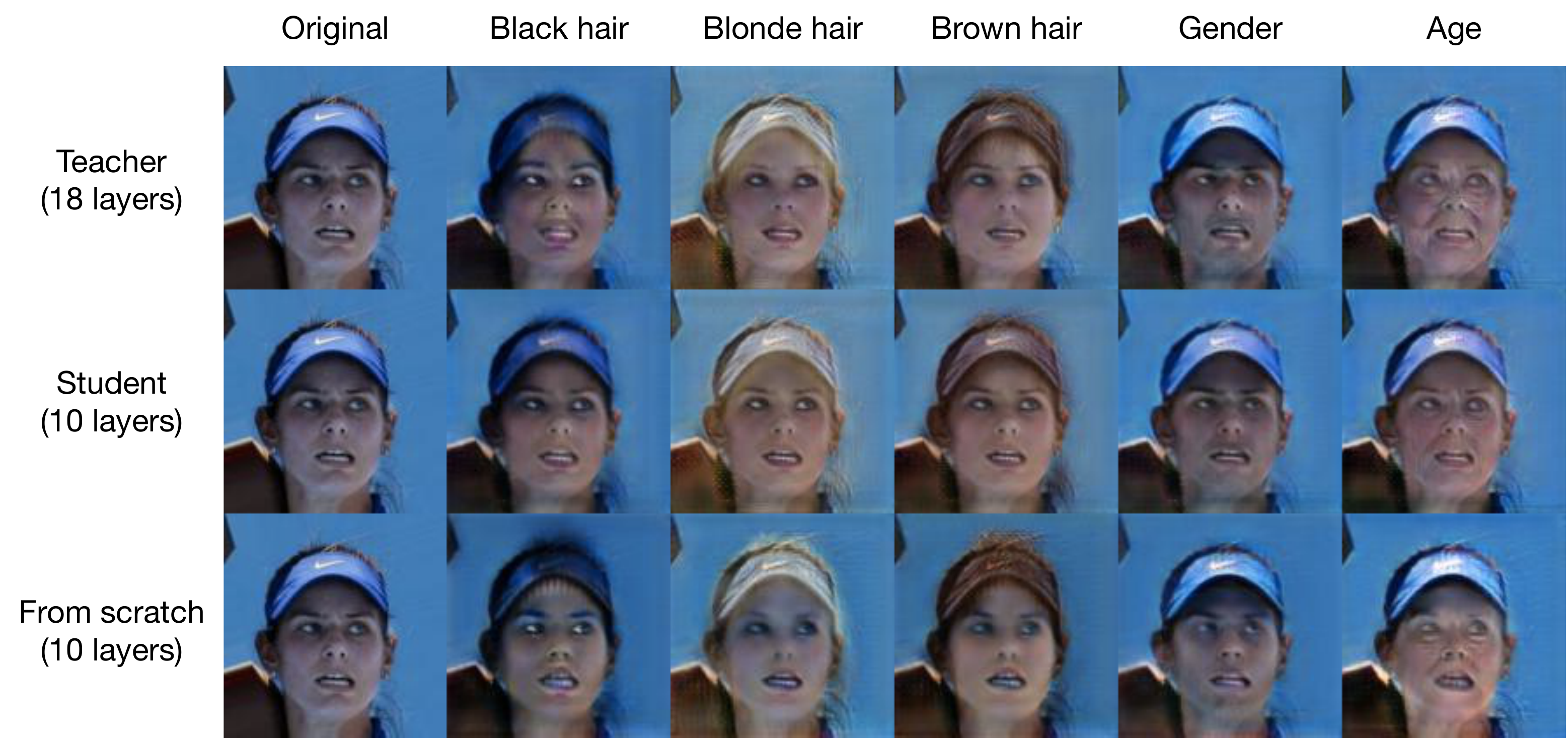}
  \vspace{-0.5em}
  \caption{Selected images from the teacher (six residual blocks), student (two
  residual blocks), and trained from scratch (two residual blocks) StarGANs. As
  shown (column two, four), LIT can appear to improve GAN performance while
  significantly compressing models. We show a randomly selected set of images in
  the Appendix. Best viewed in color.}
  \label{fig:stargan}
  \vspace{-0.5em}
\end{figure}

\minihead{LIT can compress GANs}
To the best of our knowledge, LIT is the first student/teacher method to
compress a GAN's generator, as KD is not applicable to the pixels (i.e., not
probabilities) that GAN generators output.

\begin{table}[t!]
\centering
\small
\setlength\itemsep{2em}
\begin{tabular}{lcc}
  Model               & Inception score (higher is better) & FID score (lower is better) \\ \hline
  Teacher (18 layers)              & 3.49          & 6.43\\
  LIT student (10 layers)          & \textbf{3.56} & \textbf{5.84} \\
  Trained from scratch (10 layers) & 3.37          & 6.56 \\
  Randomly initialized (10 layers) & 2.63          & 94.00 \\
  Randomly initialized (18 layers) & 2.45          & 151.43
\end{tabular}
\vspace{-0.5em}
\caption{Inception and FID scores for different versions of StarGAN. Despite
having fewer layers than the teacher, the LIT student model achieves the best
scores.}
\label{table:inception-score}
% \vspace{-0.5em}
\end{table}

We compressed StarGAN's generator~\citep{choi2017stargan} using the LIT
procedure with $\beta = 0$ (i.e., only using the intermediate representation
loss). The original StarGAN has 18 total convolutional layers (including
transposed convolutional layers), with 12 of the layers in the residual blocks
(for a total of six residual blocks). We compressed the six residual blocks to
two residual blocks (i.e., 12 to four layers) while keeping the rest of the
layers fixed. The remaining layers for the teacher model were copied to the
student model and fine-tuned. The discriminator remained fixed.

As shown in Table~\ref{table:inception-score}, LIT outperforms all baselines in
inception and FID score. Additionally, as shown in Figure~\ref{fig:stargan}, the student
model appears to perceptually outperform both the teacher model and equivalent
model trained from scratch, suggesting LIT can both compress GANs and serve as a
form of regularization.

\subsection{Impact of Training Techniques}
\label{sec:eval-training}

\begin{table}[t!]
\centering
\small
\setlength\itemsep{2em}
\begin{tabular}{ll}
  Type  & Accuracy \\ \hline
  LIT   & \textbf{93.25\%} \\
  KD    & 92.75\% \\
  One IR, teacher input              & 92.74\% \\
  One IR, no teacher input (FitNets) & 92.68\% \\
  Multiple IRs, no teacher input     & 90.42\% \\
  % LIT (splits between blocks) & 91.15\%
\end{tabular}
\begin{tabular}{ll}
  Type  & Accuracy \\ \hline
  LIT   & \textbf{94.72\%} \\
  KD    & 94.42\% \\
  One IR, teacher input              & 94.21\% \\
  One IR, no teacher input (FitNets) & 94.18\% \\
  Multiple IRs, no teacher input     & 91.27\%
\end{tabular}
\vspace{-0.5em}
\caption{Ablation study of LIT. We performed LIT, KD, and three modifications of
LIT. As shown, LIT outperforms KD and the modifications, while all the
modifications underperform standard KD. \textbf{Left:} ResNet, \textbf{Right:}
ResNeXt.}
\label{table:ablation-ir}
\vspace{-0.7em}
\end{table}

LIT uses block-wise training with the teacher IRs as input to the student
model. To show the effectiveness of block-wise training, we tried other
variations: 1) matching a single IR, with no input from the teacher (i.e.,
standard hint training/FitNets), 2) a single IR with teacher input, 3) multiple
IRs with no teacher input. We performed these variations on a teacher model of
ResNet-110 and a student model of ResNet-20 on CIFAR10 and similarly for ResNeXt.

As shown in Table~\ref{table:ablation-ir}, none of the three variants are as
effective as LIT or KD. Thus, we see that LIT's block-wise training is critical
for high accuracy compression.

\subsection{LIT is complementary to pruning}
Pruning is a key technique in deep compression in which parts of a network are
set to zero, which reduces the number of weights and, on specialized hardware,
reduces the computational footprint of networks. To see if LIT models are
amenable to pruning, we pruned ResNets trained via LIT. We additionally pruned
ResNets trained from scratch.

\begin{figure}[t!]
  \centering
  \includegraphics{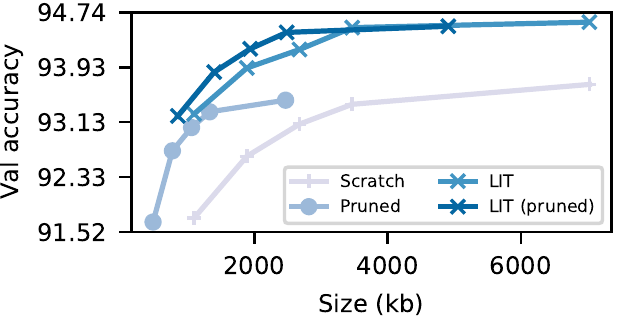}
  \vspace{-0.5em}
  \caption{The size vs accuracy of various ResNets pruned on CIFAR10. LIT is
  pareto optimal for model size and accuracy.}
  \label{fig:compression}
  \vspace{-0.7em}
\end{figure}

As shown in Figure~\ref{fig:compression}, LIT models are pareto optimal in
accuracy vs model size. Additionally, LIT models can be pruned, although less
than their trained-from-scratch counterparts. However, LIT models are more
accurate and are thus likely learning more meaningful representations. Thus, we
expect LIT models to be more difficult to prune, as each weight is more
important.

\subsection{Sensitivity Analysis of Hyperparameters}
\label{sec:hyperparameters}

% We selected the hyperparameters as described in Section~\ref{sec:methods}. We
% perform a sensitivity analysis of the hyperparameters to validate our choices.

\begin{table}[t!]
\centering
\small
\setlength\itemsep{2em}
\begin{tabular}{lcc}
  Model   & Loss        & Accuracy \\ \hline
  ResNet  & L2          & $93.20 \pm 0.04$ \\
  ResNet  & L1          & $93.19 \pm 0.05$ \\
  ResNet  & Smoothed L1 & $93.02 \pm 0.06$
\end{tabular}
\quad
\begin{tabular}{lcc}
  Model   & Loss        & Accuracy \\ \hline
  ResNeXt & L2          & $94.63 \pm 0.07$ \\
  ResNeXt & L1          & $94.62 \pm 0.07$ \\
  ResNeXt & Smoothed L1 & $93.86 \pm 0.08$
\end{tabular}
\vspace{-0.5em}
\caption{Effect of intermediate representation loss on student model accuracy.
L2 and L1 do not significantly differ, but smoothed L1 degrades accuracy.
Average of three runs.}
\label{table:penalty}
\vspace{-0.7em}
\end{table}

\minihead{Intermediate loss penalty}
To see the affect of the intermediate loss penalty, we performed LIT from a
teacher model of ResNet-110 to a student of ResNet-20 with the L1, L2, and
smoothed L1 loss. The results are shown in Table~\ref{table:penalty}. As shown,
L2 and L1 do not significantly differ, but smoothed L1 degrades accuracy.

\begin{figure}[t!]
  \centering
  \begin{subfigure}{.49\columnwidth}
    \includegraphics{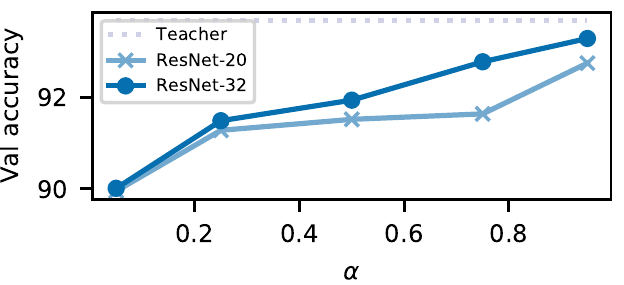}
    \vspace{-0.5em}
    \caption{CIFAR10, ResNet, $\alpha$ sensitivity}
  \end{subfigure}
  \begin{subfigure}{.49\columnwidth}
    \includegraphics{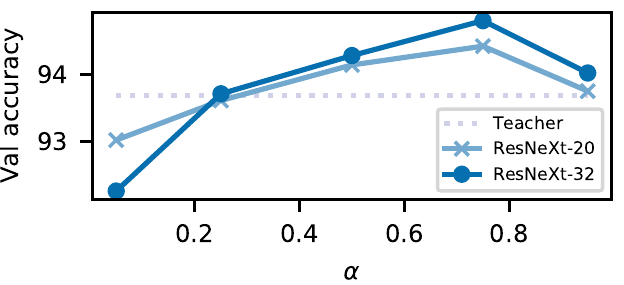}
    \vspace{-0.5em}
    \caption{CIFAR10, ResNeXt, $\alpha$ sensitivity}
  \end{subfigure}
  \vspace{-0.5em}
  \caption{The accuracy of student models as $\alpha$ (KD's interpolation factor
  for the cross-entropy and logit loss) varies for ResNet and ResNeXt on
  CIFAR10. The optimal $\alpha$ varies by model type.}
  \label{fig:alpha-cifar10}
  \vspace{-0.7em}
\end{figure}

\begin{figure}[t!]
  \centering
  \begin{subfigure}{.49\columnwidth}
    \includegraphics{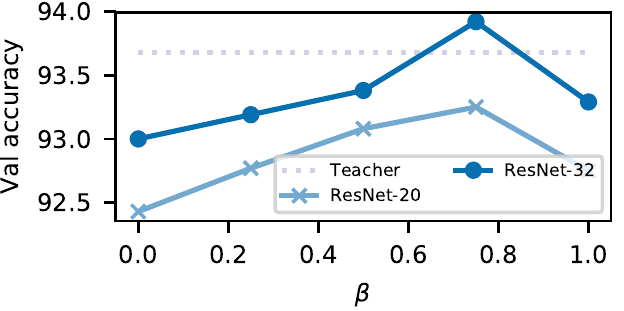}
    \vspace{-0.5em}
    \caption{CIFAR10, ResNet, $\beta$ sensitivity}
  \end{subfigure}
  \begin{subfigure}{.49\columnwidth}
    \includegraphics{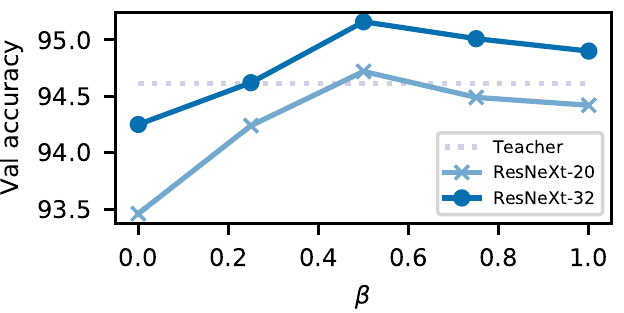}
    \vspace{-0.5em}
    \caption{CIFAR10, ResNeXt, $\beta$ sensitivity}
  \end{subfigure}
  \vspace{-0.5em}
  \caption{The accuracy of student models as $\beta$ (LIT's interpolation factor
  between KD loss and IR loss) varies for ResNet and
  ResNeXt on CIFAR10. As shown, LIT outperforms training only via KD ($\beta =
  1$) and only via intermediate representations ($\beta = 0$). The optimal
  $\beta$ appears to be lower (i.e., closer to only using the intermediate
  representation loss) for more accurate models; we hypothesize that more
  accurate models learn more informative intermediate representations, which
  helps the students learn better.}
  \label{fig:beta-cifar10}
  \vspace{-0.7em}
\end{figure}

\minihead{$\alpha$ and $\beta$}
Recall that $\alpha$ is the weighting parameter in KD and $\beta$ is the
relative weight of KD vs the intermediate representation loss
(Section~\ref{sec:methods}).

To see the effect of of $\alpha$ (which is a KD hyperparameter), we varied
$\alpha$ between 0 and 1 for ResNet and ResNeXt on CIFAR10 and CIFAR100. As
shown in Figure~\ref{fig:alpha-cifar10}, $\alpha$ can significantly affect
accuracy. Thus, we searched for $\alpha$ as opposed to using a static policy of 0.5 as
in~\cite{hinton2015distilling}.

We additionally varied $\beta$ between 0 and 1 for ResNet and
ResNeXt on CIFAR10. As shown in Figure~\ref{fig:beta-cifar10}, the optimal
$\beta$ varies between architectures but appears to be consistent within the
same meta-architecture.

\begin{table}[t!]
\centering
\small
\setlength\itemsep{2em}
\begin{tabular}{llc}
  Model  & Precision & Accuracy \\ \hline
  ResNet & fp32      & $93.20 \pm 0.04$ \\
  ResNet & Mixed     & $93.17 \pm 0.07$
\end{tabular}
\quad
\begin{tabular}{llc}
  Model   & Precision & Accuracy \\ \hline
  ResNeXt & fp32      & $94.63 \pm 0.07$ \\
  ResNeXt & Mixed     & $94.57 \pm 0.10$
\end{tabular}
\vspace{-0.7em}
\caption{Affect of mixed-precision training on the LIT procedure.
Mixed-precision training does not significantly affect the accuracy of the LIT
procedure. Average of three runs.}
\label{table:mixed-precision}
\vspace{-0.7em}
\end{table}

\minihead{LIT works with mixed precision}
To confirm mixed precision training~\citep{micikevicius2017mixed} works with
LIT, we ran LIT on ResNet and ResNeXt (the teacher
had 110 layers and the student had 20 layers) on CIFAR10 with both fp32 and
mixed precision training. The results are shown in
Table~\ref{table:mixed-precision}. As shown, mixed precision results in a
limited difference (0.06\%) in accuracies for both ResNet and ResNeXt.

\section{Conclusion}

We introduce LIT, a novel model compression technique that trains a
student model from a teacher model's intermediate representations. LIT
requires at least one intermediate layer of the student and teacher to
match in width, which allows parts of the teacher model to be copied
to the student model. By combining several such intermediate layers,
LIT students learn a high quality representation of the teacher state
without the associated depth. To overcome the lack of useful
intermediate representations within the student model at the beginning
of training, LIT uses the teacher's intermediate representations as
input to the student model during training. We show that LIT can
compress models up to 5.5$\times$ with no loss in accuracy on standard
classification benchmark tasks, outperforming standard KD and
hint training. We also show that LIT can compress StarGAN's generator, which
is, to our knowledge, the first time a student/teacher training
technique has been used to compress a GAN's generator.

\subsubsection*{Acknowledgments}
% \section*{Acknowledgments}
This research was supported in part by affiliate members and other supporters of
the Stanford DAWN project---Ant Financial, Facebook, Google, Intel, Microsoft,
NEC, Teradata, SAP, and VMware---as well as DARPA under No. FA8750-17-2-0095
(D3M), industrial gifts and support from Toyota Research Institute, Keysight
Technologies, Hitachi, Northrop Grumman, NetApp, and the NSF under grants
DGE-1656518 and CNS-1651570.

\bibliography{lit-iclr}
\bibliographystyle{iclr2019_conference}

\pagebreak
\begin{appendix}

\section{Experimental Details}

All network architectures we use are standard architectures for the datasets of
choice. All student/teacher pairs were of the same architecture type (e.g.,
ResNet to ResNet). For each dataset, we detail the architecture and
hyperparameters used.

All experiments were performed in PyTorch v0.4.0 with Python 3.5. Experiments
were run on public cloud and custom servers using NVIDIA P100, V100, Titan Xp,
and Titan V GPUs.

All weights for architectures were initialized as in the original architecture.

\subsection{CIFAR10/100}

We used the same hyperparameters for CIFAR10 and CIFAR100.

\minihead{ResNet}
We use standard ResNets~\citep{he2016deep} for
CIFAR10/100~\citep{krizhevsky2009learning,}. The architectures are parameterized
by the number of ``computation heavy" layers, i.e., convolutional and fully
connected layers, but not batch norm and ReLU layers. Thus, a ResNet-110 has
110 convolutional and fully connected layers.

Each ResNet has three blocks, where the convolutional layers in each block have
the same number of filters. The last convolutional layer in each block
downsamples by a factor of two. The ResNets can be parameterized by the number
of residual blocks in each block, where each residual block has two
convolutional layers. This parameterization is the same parameterizing by the
number of layers. For example, an [18, 18, 18] has 110 layers total, 108 layers
in the blocks, along with an additional convolutional layer at the start and a
fully connected layer. We show a table of the number of layers and block
parameterizations in Table~\ref{table:resnet-types}.

\begin{table}
\centering
\setlength\itemsep{2em}
\begin{tabular}{ll}
  Model      & Parameterization \\ \hline
  ResNe(X)t-20  & [3, 3, 3] \\
  ResNe(X)t-32  & [5, 5, 5] \\
  ResNe(X)t-44  & [7, 7, 7] \\
  ResNe(X)t-56  & [9, 9, 9] \\
  ResNe(X)t-110 & [18, 18, 18]
\end{tabular}
\caption{ResNet and ResNeXt types for CIFAR10/100.}
\label{table:resnet-types}
\end{table}

For all experiments, we used a batch size of 32, SGD with a momentum of
0.9, and weight decay of 1e-4.

For training from scratch, we trained with a starting learning rate of 0.1 for
200 epochs with milestones at 100 and 150 epochs, decaying the learning rate by
a factor of 10.

For KD, we trained with a starting learning rate of 0.1 for 250 epochs with
milestones at 100 and 175 epochs, decaying the learning rate by a factor of 10.
We used $\tau = 6$ and $\alpha = 0.95$.

For LIT, we trained with a starting learning rate of 0.1 for 175 epochs with
milestones at 60, 100, and 125 epochs, decaying the learning rate by a factor of
10. We then fine-tuned using the KD loss for another 75 epochs with a starting
learning rate of 0.01 and milestones at 35 and 55 epochs. We used $\beta =
0.75$.

\minihead{ResNeXt}
We use standard ResNeXts~\citep{xie2017aggregated} for CIFAR10/100. The
parameterization in terms of number of layers are the same for ResNet.

ResNeXt has an additional parameter of the group cardinality. We use a group
cardinality of 32 for all experiments, except the ones detailed below.

For all experiments, we used a batch size of 32, SGD with a momentum of 0.9, and
weight decay of 1e-4.

For training from scratch, we trained with a starting learning rate of 0.1 for
300 epochs with milestones at 150 and 225 epochs, decaying the learning rate by
a factor of 10.

For KD, we trained with a starting learning rate of 0.1 for 300 epochs with
milestones at 100, 175, and 225 epochs, decaying the learning rate by a factor
of 10. We used $\tau = 6$ and $\alpha = 0.95$.

For LIT, we trained with a starting learning rate of 0.1 for 200 epochs with
milestones at 100, 145, and 175 epochs, decaying the learning rate by a factor
of 10. We then fine-tuned using the KD loss for another 125 epochs with a
starting learning rate of 0.01 and milestones at 65, 95, and 110 epochs. We used
$\beta = 0.5$.

\minihead{Reduced cardinality ResNeXt}

All hyperparameters were the same as the standard ResNeXt experiments except we
used $\beta = 0.25$ and a student cardinality of 16.

\subsection{Amazon Reviews}

We use standard VDCNNs~\citep{conneau2016very} for Amazon Reviews full and
polarity~\citep{he2016ups}. VDCNN has an initial convolutional layer and four
blocks of convolutional layers (each block has the same number, but vary between
types of VDCNNs). Thus, a VDCNN-9 has an initial convolutional layer and two
convolutional layers in each subsequent block. We consider VDCNN-9, VDCNN-17,
and VDCNN-29 as in~\cite{conneau2016very}.

For all experiments, we used a batch size of 128, SGD with a momentum of 0.9,
and weight decay of 1e-4.

For training from scratch, we trained with a starting learning rate of 0.01 for
15 epochs, with milestones at 3, 6, 9, 12, and 15 epochs, decaying the learning
rate by a factor of 10.

For KD, we trained with a starting learning rate of 0.01 for 18 epochs, with
milestones at 3, 6, 9, 12, and 15 epochs, decaying the learning rate by a factor
of 10. We used $\tau = 6$ and $\alpha = 0.98$.

For LIT, we trained with a starting learning rate of 0.01 for 15 epochs, with
milestones at 3, 6, 9, and 12 epochs, decaying the learning rate by a factor of
2. We then fine-tuned using the KD loss for another 10 epochs with a starting
learning rate of 0.000625 and milestones at 4 and 8 epochs. We used $\beta =
0.02$.

\subsection{StarGAN}

We used the StarGAN as described in~\cite{choi2017stargan}. The original StarGAN
has 18 total convolutional layers (including transposed convolutional layers),
with 12 of the layers in the residual blocks (each residual block has two
convolutional layers). We compressed the six residual blocks to two residual
blocks.

For all experiments, we used a batch size of 16, SGD with a momentum of 0.9, and
weight decay of 1e-4.

For training from scratch, we trained with a starting learning rate of 0.0001
for 20 epochs. The learning rate was decayed to 0 over the last 10 epochs.

As KD does not apply, we did not run KD.

For LIT, we trained with a starting learning rate of 0.0001 for 16 epochs,
decaying the learning rate by 10 at epoch 8 (only the IR loss was used). We then
fine-tuned with the discriminator with a starting learning rate of 0.00005 for
10 epochs, decaying the learning rate by 10 at epoch 5.

\begin{figure}[t!]
  \centering
  \begin{subfigure}{0.32\columnwidth}
    \includegraphics[width=0.99\columnwidth]{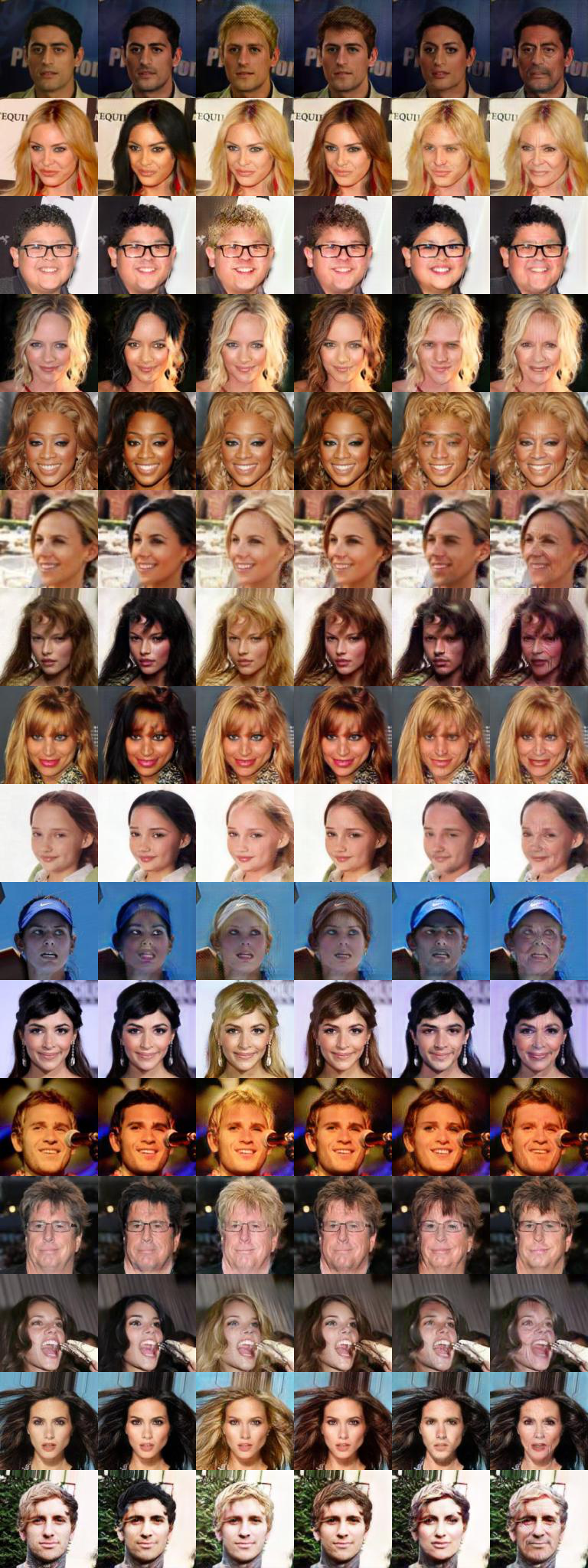}
    \caption{Teacher (18 layers)}
  \end{subfigure}
  \begin{subfigure}{0.32\columnwidth}
    \includegraphics[width=0.99\columnwidth]{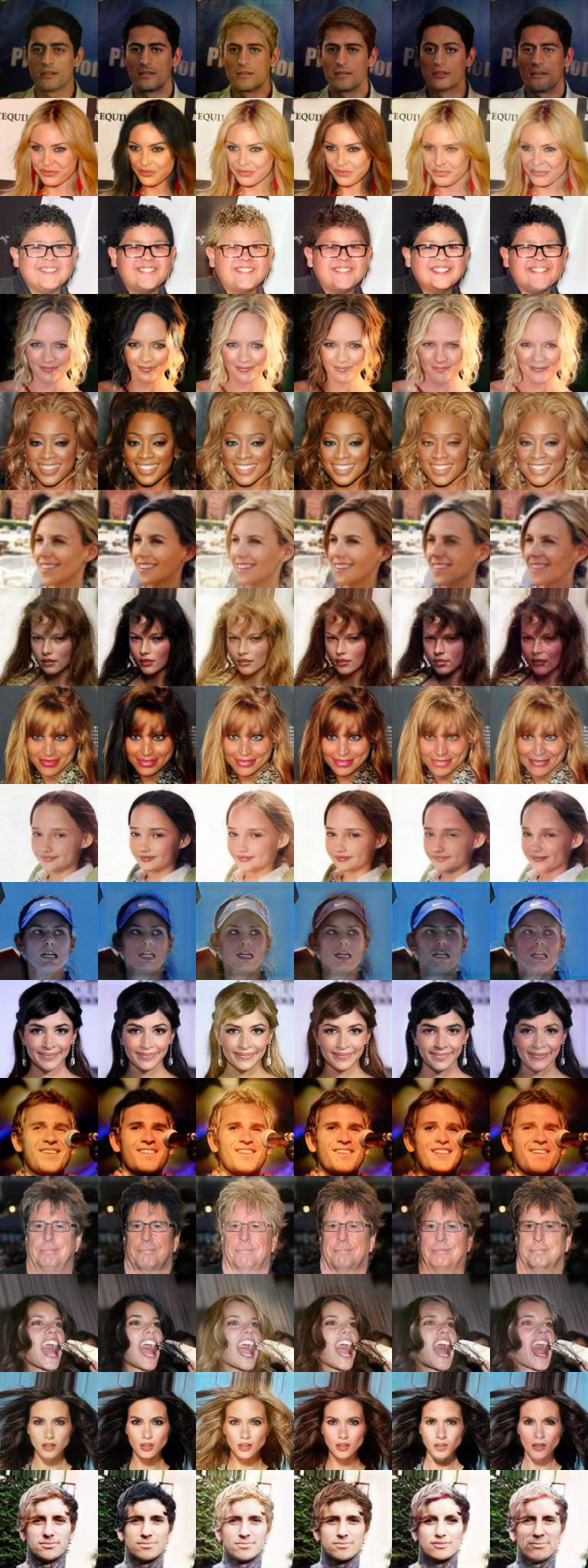}
    \caption{Student (10 layers)}
  \end{subfigure}
  \begin{subfigure}{0.32\columnwidth}
    \includegraphics[width=0.99\columnwidth]{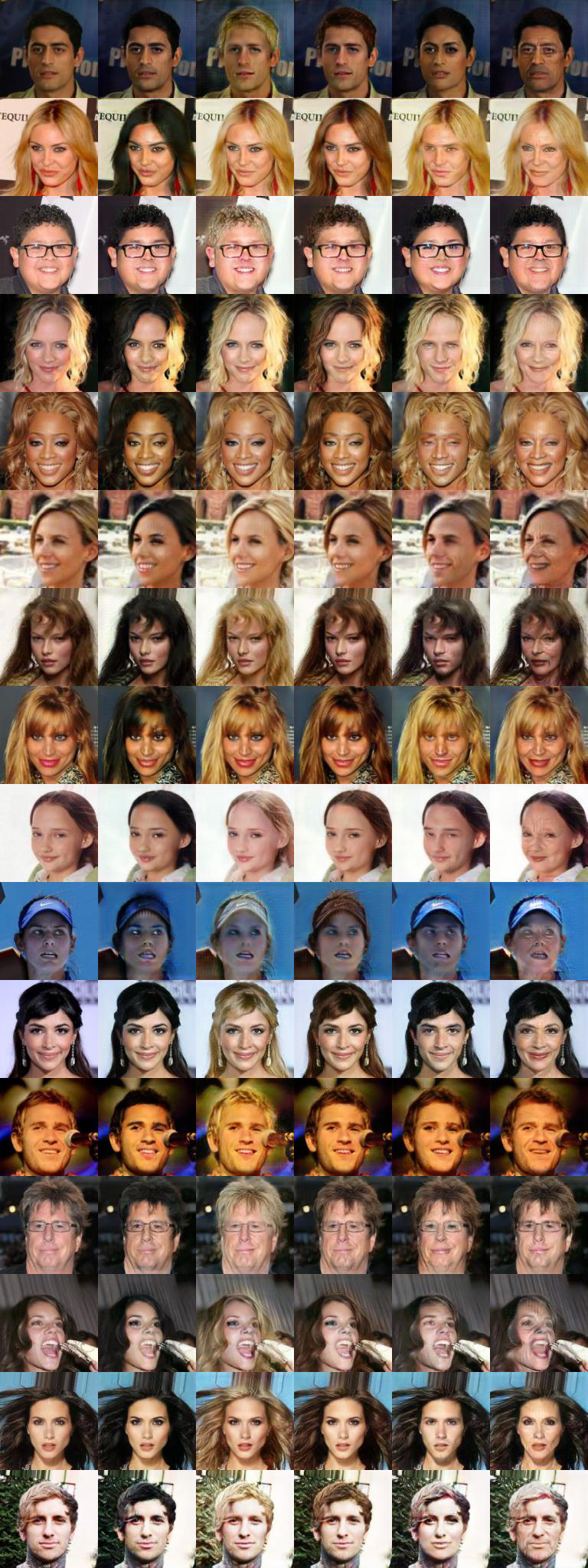}
    \caption{Trained from scratch (10 layers)}
  \end{subfigure}
  \caption{Randomly selected images from the teacher (six residual blocks, 18
  total layers),
  student (two residual blocks, 10 total layers), and trained from scratch (two
  residual blocks, 10 total layers)
  StarGANs. As shown, LIT can appear to improve GAN performance while
  significantly compressing models. The columns are: Original, Black Hair, Blond Hair,
  Brown Hair, Male, Age.}
  \label{fig:stargan-many}
\end{figure}

\section{StarGAN images}

We show a randomly selected set of images generated from the StarGAN teacher,
student, and trained from scratch generators in Figure~\ref{fig:stargan-many}.

\end{appendix}

\end{document}